\documentclass[12pt]{article}

\usepackage{sbc-template}

\usepackage{graphicx,url}

\usepackage[utf8]{inputenc}

\title{Avaliação de eficiência na leitura: uma abordagem \\baseada em PLN}

\author{Túlio Sousa de Gois\inst{1,2}, Raquel Meister Ko. Freitag\inst{1,3}}

\address{Laboratório Multiusuário de Informática e Documentação Linguística \\ Universidade Federal de Sergipe -- Brazil (UFS)\\
Didática II -- 49.107-230 -- São Cristóvão -- SE -- Brazil
\nextinstitute
  Departamento de Computação -- Universidade Federal de Sergipe -- UFS
\nextinstitute
  Departamento de Letras Vernáculas -- Universidade Federal de Sergipe -- UFS
  \email{\{tuliosg, rkofreitag\}@academico.ufs.br}
}

\begin{document} 

\maketitle

\begin{abstract}
  The cloze test, widely used due to its low cost and flexibility, makes it possible to assess reading comprehension by filling in gaps in texts, requiring the mobilization of diverse linguistic repertoires. However, traditional correction methods, based only on exact answers, limit the identification of nuances in student performance. This study proposes an automated evaluation model for the cloze test in Brazilian Portuguese, integrating orthographic (edit distance), grammatical (\textit{POS tagging}) and semantic (similarity between \textit{embeddings}) analyses. The integrated method demonstrated its effectiveness, achieving a high correlation with human evaluation ($\rho = 0.832$). The results indicate that the automated approach is robust, sensitive to variations in linguistic repertoire and suitable for educational contexts that require scalability.

\end{abstract}
     
\begin{resumo} 
  O teste cloze, amplamente difundido por seu baixo custo e flexibilidade, permite avaliar a compreensão leitora por meio do preenchimento de lacunas em textos, exigindo mobilização de repertórios linguísticos diversos. No entanto, os métodos tradicionais de correção, baseados apenas em respostas exatas, limitam a identificação de nuances no desempenho dos estudantes. Este estudo propõe um modelo automatizado de avaliação para o teste cloze em português brasileiro, integrando análises ortográfica (distância de edição), gramatical (via \textit{POS tagging}) e semântica (similaridade entre \textit{embeddings}). O método integrado mostrou-se efetivo, atingindo uma alta correlação com a correção humana ($\rho = 0{,}832$). Os resultados indicam que a abordagem auto\-matizada é robusta, sensível às variações do repertório linguístico e adequada para contextos educacionais que exigem escalabilidade. 
\end{resumo}

\section{Introdução}

Os problemas de leitura no Brasil são historicamente graves e foram intensificados pela pandemia de COVID-19, que expôs e acirrou ainda mais as desigualdades e educacionais preexistentes. Dados do PISA 2022 revelam que os estudantes brasileiros continuam a\-presentando desempenho insatisfatório em leitura, com escore de 410 pontos, inferior à de países vizinhos como Chile e Uruguai. Esses resultados reiteram que a leitura, enquanto ferramenta básica para a construção do conhecimento e da cidadania, não tem sido desenvolvida de maneira efetiva na escola, comprometendo a formação crítica dos estudantes e sua participação plena na sociedade.

As avaliações diagnósticas, como Prova Brasil, PISA e PIRS, apresentam o panorama do sistema, mas não permitem que sejam implementadas medidas específicas em tempo hábil para atenuar as dificuldades dos estudantes com a leitura, de modo que as intervenções pedagógicas adequadas possam ser realizadas ainda no mesmo ano letivo. Assim, no desenvolvimento de instrumentos e matrizes de avaliação para o acompa\-nhamento próximo e sistemático da leitura, o teste cloze tem se mostrado promissor por sua ampla difusão e baixo custo operacional \cite{freitas2025use}.

O teste cloze é um instrumento que avalia a compreensão leitora a partir da omissão de palavras em um texto, que devem ser completadas pelo estudante com base em seus conhecimentos linguísticos, textuais e contextuais. Por exigir a mobilização de diferentes habilidades cognitivas e linguísticas, o teste cloze permite diagnosticar não apenas o vocabulário e a gramática dominados pelo estudante, mas também seu grau de proficiência em leitura. Além de ser autoaplicável, esse tipo de teste pode ser adaptado a diferentes níveis de ensino e utilizado de forma contínua para acompanhar o desenvolvimento da leitura, tornando-se uma estratégia eficaz para planejar intervenções e promover avanços concretos no processo de aprendizagem.

O processo de correção do teste, no entanto, ainda baseia-se em um modelo de respostas certas/erradas, o que restringe a potencialidade de identificar especificidades da deficiência na compreensão leitora, em especial a diferenciação entre os níveis do repertório do estudante e o seu conhecimento gramatical \cite{cardoso2024eficiencia}. O uso de ferramentas de processamento de linguagem natural tem permitido avançar neste campo, com a adoção de procedimentos para avaliação automática que saiam do limite do certo e errado, utilizando, por exemplo, similaridade semântica para acessar as lacunas preenchidas \cite{de2024nlp}.

Na continuidade do desenvolvimento e validação de um método automatizado para a avaliação de testes cloze em português brasileiro, expandindo os critérios de correção para além da tradicional verificação de respostas exatas, neste trabalho visamos integrar a análise ortográfica (via distância de edição), gramatical (via \textit{POS tagging}) e semântica (via similaridade semântica de \textit{embeddings}). Ampliando a exploração de aspectos do repertório, o trabalho também compara modelos de linguagem e a arquitetura de \textit{embedding} (contextual vs. não contextual) mais eficazes para a tarefa de avaliação semântica. O desempenho do sistema final de avaliação automática através de seus resultados, é, então, comparado com as anotações de uma juíza especialista.

\section{Antecedentes} \label{sec:firstpage}
Tradicionalmente, a correção do teste cloze prioriza a resposta exata como critério de avaliação \cite{freitas2025use}, considerando acertos apenas as palavras idênticas às do texto original. No entanto, há métodos alternativos de correção e/ou aplicação de técnicas computacionais ao processo.

\cite{kleijn2019cloze} apresentam uma alternativa ao TC tradicional, o \textit{Hybrid Text Comprehension cloze (HyTeC-cloze)}. Dentre as mudanças propostas, está a correção dos testes utilizando uma pontuação semântica (também chamada de correção por resposta aceitável), que considera tanto as respostas exatas quanto as com semântica correta.  

Em uma via distinta, \cite{mirault2021algorithm} propôs um algoritmo para correção do cloze baseado em regras. O método verifica se a resposta é uma palavra válida (usando bases lexicais), aplica remoção de afixos e calcula distâncias ortográficas (via \texttt{stringdist} \cite{van2014stringdist}) para tolerar erros de grafia. Apesar do método automatizado, a abordagem não suporta a correção por resposta aceitável.

Uma estratégia baseada em categorização foi proposta por \cite{cardoso2024eficiencia}, que classificou as respostas por classe gramatical e campo semântico, considerando corretas aquelas alinhadas à palavra esperada em ambos os critérios. A eficiência em leitura foi medida combinando taxa de acerto e tempo de resolução, classificando os participantes quanto ao perfil na escala de \cite{bormuth1968cloze}. No entanto, a metodologia exige anotação manual das categorias, inviabilizando a aplicação em larga escala.

Buscando automatizar a análise semântica, o estudo de \cite{de2024nlp} utilizou \textit{word embeddings} (WEs) e similaridade de cosseno para avaliar respostas semanticamente próximas às esperadas. O ranking das palavras, gerado pela avaliação de similaridade, foi validado contra avaliações humanas. Embora promissor, o trabalho não integra critérios gramaticais ou tolerância a erros ortográficos, limitando sua aplicação prática.

Considerando estes antecedentes, neste trabalho propomos uma abordagem baseada em PLN que integra três níveis de análise: (1) verificação ortográfica via distância de Damerau-Levenshtein, (2) validação gramatical através da comparação de \textit{POS tags}, e (3) avaliação semântica das respostas. Diferentemente dos apresentados anteriormente, o método proposto implementa um sistema de pontuação que preserva a robustez contra erros ortográficos, automatiza a avaliação semântica, considera acertos de classe gramatical e contabiliza as respostas por tipo.

\section{Método}

\subsection{Dados}
Os dados utilizados neste trabalho são provenientes da aplicação de testes cloze que ocorreu no Colégio de Aplicação da Universidade Federal de Sergipe (CODAP/UFS). Na coleta, foram aplicados quatro testes diferentes, em turmas do 6º ao 9º ano do ensino fundamental ($n=210$) \cite{santos2025aspecto}.

As respostas foram tabuladas em planilhas, onde cada linha representa uma res\-posta ao teste, contendo as palavras inseridas nas lacunas e as informações do aluno respondente. Após essa etapa, os testes foram corrigidos por uma juíza especialista seguindo o padrão descrito em \cite{cardoso2024eficiencia}: a resposta da lacuna era classificada quanto à classe gramatical e ao campo semântico. Para viabilizar uma análise quantitativa, foi realizado o mapeamento das categorias atribuídas pela juíza para um sistema de classificação padronizado.

\subsection{Distância de edição Damerau-Levenshtein}
A conferência de erros ortográficos é baseada no trabalho de \cite{mirault2021algorithm}, que utilizou \textit{Optimal String Alignment} (OSA) como parâmetro para aceitação da resposta. O OSA é uma variação da distância de Levenshtein, quantificando a diferença entre duas palavras pelo número mínimo de edições para transformar uma na outra, mas considerando a transposição de letras vizinhas como custo 1. Em \cite{mirault2021algorithm}, uma distância menor que 3 do gabarito era aceita. Contudo, esse valor arbitrário é pro\-blemático, pois o tamanho das palavras-alvo em um teste cloze varia.

Nossa abordagem utiliza a função \texttt{edit\_distance} da biblioteca \texttt{NLTK} \cite{bird-loper-2004-nltk}, com parâmetro para considerar transposições, que implementa a distância de Damerau-Levenshtein \cite{damerau1964technique, levenshtein1966binary}. O critério de aceitação é dinâmico: a distância deve ser menor que $\frac{1}{3}$ do tamanho da resposta espe\-rada (sendo $1$ o valor mínimo), considerando assim a quantidade de caracteres da palavra-alvo na comparação, que ocorre exclusivamente com o gabarito.

\subsection{\textit{POS tags}}
A verificação da classe gramatical de uma resposta é um indicador importante da compreensão sobre a estrutura sintática da frase \cite{cardoso2024eficiencia}, então também foi acrescida como ponto de avaliação. Para extrair essa informação, foi utilizada a biblioteca \texttt{spacy} \cite{Honnibal_spaCy_Industrial-strength_Natural_2020} e a pipeline \textit{pt\_core\_news\_lg}\footnote{\texttt{https://spacy.io/models/pt\#pt\_core\_news\_lg}}. A escolha do modelo ocorreu através devido ao desempenho na extração de \textit{POS tags} para o Português Brasileiro (acurácia de $0{,}97$)\footnote{\texttt{https://spacy.io/models/pt\#pt\_core\_news\_lg-accuracy}} e também pelo tempo de execução. 

Para alinhar o nível de detalhe do modelo (que distingue \texttt{VERB} de \texttt{AUX}) com a anotação humana (que trata ambos como ``verbo"), implementou-se uma regra de mape\-amento que agrupa a predição \texttt{AUX} na categoria \texttt{VERB} para fins de avaliação, garantindo uma comparação justa.

\subsection{Similaridade semântica}
A implementação da avaliação semântica das respostas é uma abordagem que se mostrou promissora \cite{de2024nlp}. Neste trabalho, utilizamos modelos baseados em \textit{transformers}, diferente do de \cite{de2024nlp}, que se baseou em modelos estáticos. Com base nos desempenhos para o português na tarefa de similaridade semântica proposta no ASSIN \cite{real2020assin}, foram selecionados os modelos \texttt{BERTimbau Base} \cite{souza2020bertimbau} e \texttt{Albertina 100M PTBR} \cite{albertina-pt-fostering}.

Para a seleção do modelo mais eficaz na abordagem proposta, foi conduzido um experimento para determinar se embeddings contextuais (palavra-alvo inserida na frase) superavam as não contextuais (palavra-alvo isolada). O desempenho foi medido compa\-rando os \emph{scores} de similaridade de cosseno de cada abordagem com a classificação ordinal de uma juíza especialista. Para esta etapa, as anotações da juíza foram mapeadas para uma escala ordinal: respostas exatas valiam 3; aceitáveis, 2; de classe gramatical correta, 1; e incorretas ou em branco, 0. O experimento utilizou a biblioteca \texttt{transformers} \cite{DBLP:journals/corr/abs-1910-03771} para o uso dos modelos e \texttt{pytorch} \cite{DBLP:journals/corr/abs-1912-01703} para as operações de tensores.

\subsection{Fluxo de avaliação}
\begin{figure}[ht]
\centering
\includegraphics[width=.5\textwidth]{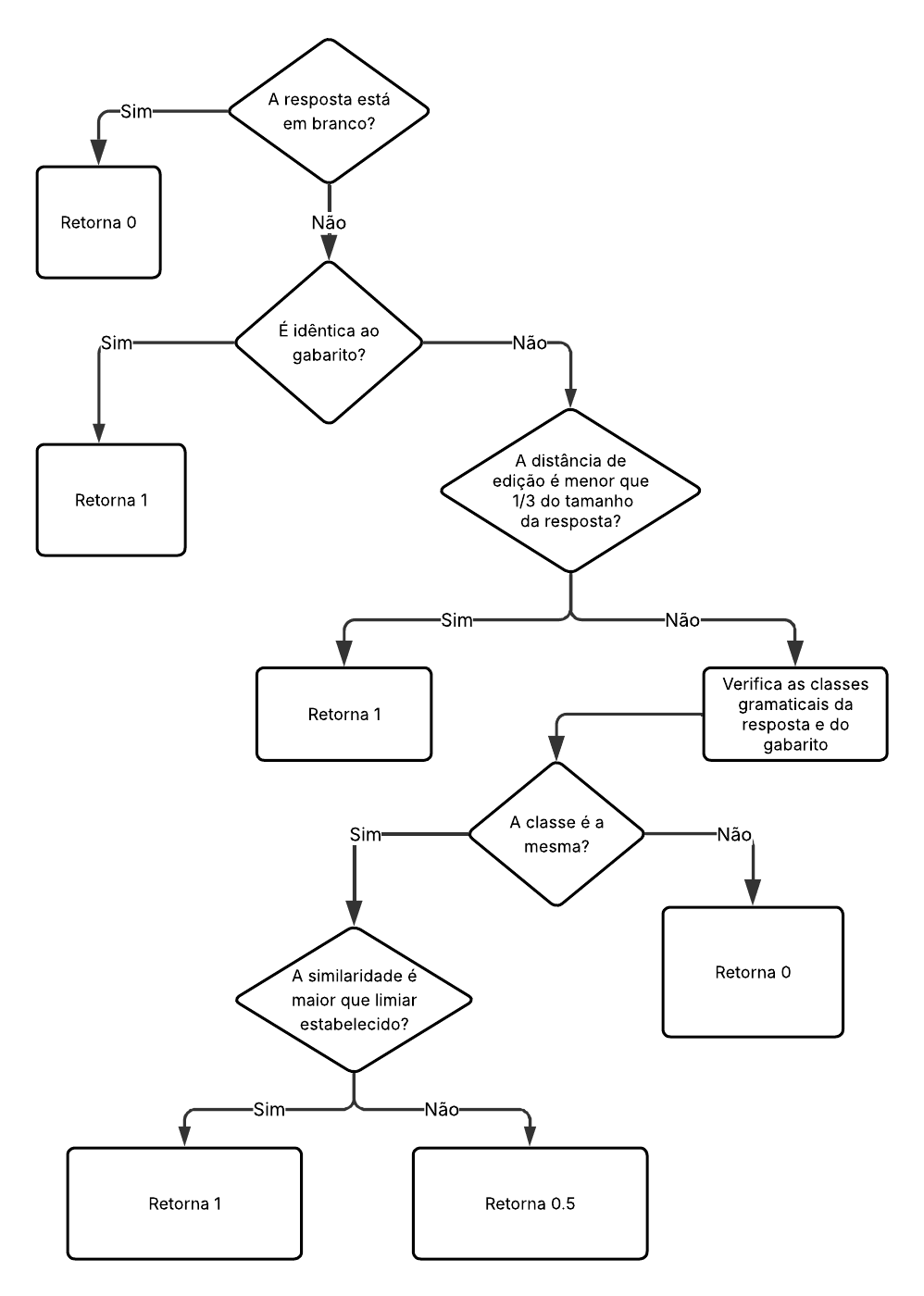}
\caption{Fluxo para correção da lacuna}
\label{fig:fluxoTC}
\end{figure}

Partindo da integração das técnicas e da validação do modelo de linguagem, o método proposto para avaliação de cada lacuna do cloze segue o fluxo hierárquico presente na Figura \ref{fig:fluxoTC}. O sistema classifica cada resposta em uma das seguintes categorias:

\begin{itemize}
\item \textbf{Exata}: idêntica ao gabarito (pontuação $1{,}0$);
\item \textbf{Grafia incorreta}: resposta com pequenos erros ortográficos em relação ao gaba\-rito (pontuação $1{,}0$);
\item \textbf{Aceitável}: resposta semanticamente equivalente ao gabarito, com mesma classe gramatical (pontuação $1{,}0$);
\item \textbf{Classe correta}: possui a mesma classe gramatical do gabarito, mas não é semanticamente equivalente (pontuação $0{,}5$);
\item \textbf{Em branco}: a não-resposta (pontuação $0$);
\item \textbf{Incorreta}: resposta que não se enquadra nas demais categorias (pontuação $0$);
\end{itemize}

A taxa de compreensão final do respondente é calculada como a média da pontuação obtida em todas as lacunas, expressa em percentual.

\subsection{Implementação}

A metodologia de avaliação descrita foi encapsulada em uma classe Python modular e reutilizável, denominada \texttt{NLPcloze}. Esta classe centraliza todas as funcionalidades necessárias, desde o carregamento dos modelos de linguagem e o gerenciamento de cache de \textit{embeddings}, até a aplicação do fluxo de avaliação hierárquico em um conjunto de dados de respostas de alunos.

A arquitetura da classe foi projetada para ser extensível, permitindo a fácil substituição de modelos de linguagem ou o ajuste de parâmetros, como o limiar de aceitação semântica. O fluxo de decisão principal, implementado no método \texttt{avaliar\_lacuna}, segue a lógica apresentada na Figura
 \ref{fig:fluxoTC}.

\section{Resultados}
\subsection{Modelo de Linguagem e Limiar de Aceitabilidade}
Os desempenhos dos modelos de linguagem foram avaliados através da correlação de Spearman com os dados ordinais da especialista, bem como por métricas de classificação (AUC e \emph{F1-score}), onde as respostas foram categorizadas entre ``aceitáveis" (\emph{scores} 2 e 3) e ``não-aceitáveis" (\emph{scores} 0 e 1).

\begin{table}[ht!]
\centering
\caption{Comparação das Métricas dos Modelos de Linguagem}
\label{tab:comparacao_modelos}
\begin{tabular}{lcccc}
\hline
\textbf{Modelo} & \textbf{AUC} & \textbf{\emph{F1-score} máximo} & \textbf{Limiar ótimo} & \textbf{Spearman} \\ \hline
BERTimbau (Contexto) & 0{,}880 & 0{,}727 & 0{,}652 & 0{,}682 \\
Albertina (Contexto) & 0{,}847 & 0{,}706 & 0{,}770 & 0{,}622 \\
BERTimbau (Palavra)  & 0{,}815 & 0{,}637 & 0{,}877 & 0{,}532 \\
Albertina (Palavra)  & 0{,}741 & 0{,}605 & 0{,}993 & 0{,}439 \\ \hline
\end{tabular}
\end{table}

Conforme exposto na Tabela \ref{tab:comparacao_modelos}, o modelo BERTimbau, utilizando contexto, a\-presentou o melhor desempenho. Ele alcançou a maior correlação de Spearman com a avaliação humana ($\rho = 0{,}682$), a maior AUC ($0{,}880$) e o maior \emph{F1-score} Máximo ($0{,}727$). 

Partindo do experimento, consideramos o ponto onde o F1 é maximizado como o limiar ótimo. Para o BERTimbau (com contexto), o limiar foi de $0{,}652$, assim, estabelecemos o ponto de ``aceitabilidade" para a nossa avaliação semântica.

\subsection{Validação da Avaliação Automática}
Após integrar o modelo e o limiar selecionados no fluxo de avaliação, a implementação final da abordagem foi validada contra o gabarito humano em duas frentes: análise de \emph{scores} e concordância de categorias.

Primeiramente, foi comparada a pontuação final gerada pelo sistema com o \emph{score} ordinal mapeado da avaliação da especialista (ver Seção 3.4). A abordagem proposta obteve uma alta correlação de Spearman, $\rho = 0{,}832$.

Em seguida, foi avaliada a concordância entre os rótulos categóricos. Para uma comparação justa, os rótulos do sistema (incluindo ``grafia incorreta") e da juíza foram padronizados em um conjunto comum de classes (exata, aceitável, classe correta, incorreta). O coeficiente Kappa de Cohen ($\kappa$) foi de $0{,}727$, indicando uma concordância substancial, de acordo com \cite{landis1977measurement}.

\section{Conclusões}

Neste estudo, expandimos os critérios tradicionais de correção em testes cloze para além da verificação de respostas exatas, ao integrar análises ortográfica (por meio da distância de edição), gramatical (via \textit{POS tagging}) e semântica (com base na similaridade de cosseno entre \textit{embeddings}). Também comparamos diferentes modelos de linguagem e arquiteturas de \textit{embedding}, contextual e não contextual, para identificar as soluções mais eficazes na avaliação automatizada de aceitabilidade semântica. O desempenho do sistema final foi confrontado com as anotações de uma juíza especialista, a fim de validar sua aderência aos critérios humanos de julgamento.

Entre os modelos avaliados, o BERTimbau com contexto se destacou como o mais eficaz. Esse modelo apresentou os melhores resultados em todas as métricas analisadas: maior correlação de Spearman com os julgamentos humanos ($\rho = 0{,}682$), maior AUC ($0{,}880$) e \emph{F1-score} máximo ($0{,}727$). Esses resultados demonstram sua capacidade de captar nuances contextuais relevantes para o julgamento semântico, o que é essencial em tarefas que demandam sensibilidade linguística próxima à humana.

Na fase de validação do método, os resultados apontaram para a alta correlação de Spearman ($\rho = 0{,}832$) entre os \emph{scores} gerados pela correção automática e os atribuídos pela especialista, sugerindo forte alinhamento entre as avaliações. Além disso, a concordância categórica aferida pelo coeficiente Kappa de Cohen ($\kappa = 0{,}727$) foi conside\-rada substancial, \cite{landis1977measurement}, evidenciando que, mesmo após a categorização dos \emph{scores}, o sistema mantém correspondência com as decisões humanas. 

Estes resultados sugerem a confiabilidade da abordagem proposta, apontando seu potencial para uso em contextos educacionais e em pesquisa que demandam escalabilidade, precisão e sensibilidade semântica na avaliação de teste cloze em larga escala. Em suma, este estudo reforça a importância de aplicações em processamento de linguagem natural para o desenvolvimento de instrumentos e matrizes de avaliação para o acompa\-nhamento próximo e sistemático da leitura, contribuindo para a educação de qualidade.

\section{Limitações e Trabalhos futuros}
Os testes cloze que resultaram nos dados utilizados no presente trabalho tinham apenas lacunas de verbos, o que pode enviesar os resultados aqui apresentados.

A necessidade de mapeamentos e a diferença entre as classificações atribuídas pela juíza especialista e pela abordagem proposta podem afetar a avaliação, a superestimando ou subestimando. 
Em trabalhos futuros, é indicada a avaliação por mais de um especialista, testes com lacunas de outras classes gramaticais e protocolos de correção bem definidos para facilitar a validação posterior.

\section*{Disponibilidade de Dados e Códigos}
O conjunto de dados anonimizado utilizado para a avaliação, bem como o código-fonte completo da classe \texttt{NLPcloze} e os notebooks utilizados para as análises apresentadas neste trabalho, estão publicamente disponíveis em um repositório no GitHub para fins de reprodutibilidade e reuso: \texttt{https://github.com/tuliosg/nlp-cloze}.

\section*{Agradecimentos}
Este trabalho está vinculado ao projeto \textit{Impactos da pandemia de COVID-19 na linguagem da criança e do adulto: foco no desenvolvimento e na aprendizagem da leitura}, financiado pelo edital Capes 12/2021 - PDPG Impactos da Pandemia. Agradecemos ao suporte de infraestrutura e  equipe do Laboratório Multiusuário de Informática e Documentação Linguística (LAMID) da Universidade Federal de Sergipe.
Agradecemos também à Rede Brasileira de Reprodutibilidade pelo suporte financeiro, permitindo a participação no STIL.

\bibliographystyle{sbc}
\bibliography{sbc-template}

\end{document}